\pdfoutput=1
\documentclass[11pt]{article}

\usepackage[final]{acl}

\usepackage{times}
\usepackage{latexsym}

\usepackage[T1]{fontenc}
\usepackage[utf8]{inputenc}

\usepackage{microtype}

\usepackage{inconsolata}

\usepackage{graphicx}
\usepackage{booktabs}
\usepackage{amsmath}
\usepackage{amsfonts}
\usepackage{threeparttable}

\title{Efficient OCR for Building a Diverse Digital History}

\author{
  \textbf{Jacob Carlson\textsuperscript{1}},
  \textbf{Tom Bryan\textsuperscript{1}},
  \textbf{Melissa Dell\textsuperscript{1,2}},
\\
  \textsuperscript{1}Harvard University, Cambridge, MA, USA\\
  \textsuperscript{2}National Bureau of Economic Research, Cambridge, MA, USA
\\
  \texttt{\{jacob\_carlson,tom\_bryan,melissadell\}@fas.harvard.edu}
}

\begin{document}
\maketitle
\begin{abstract}
Many users consult digital archives daily, but the information they can access is unrepresentative of the diversity of documentary history. The sequence-to-sequence architecture typically used for optical character recognition (OCR) -- which jointly learns a vision and language model -- is poorly extensible to low-resource document collections, as learning a language-vision model requires extensive labeled sequences and compute. This study models OCR as a character level image retrieval problem, using a contrastively trained vision encoder. Because the model only learns characters’ visual features, it is more sample efficient and extensible than existing architectures, enabling accurate OCR in settings where existing solutions fail. Crucially, it opens new avenues for community engagement in making digital history more representative of documentary history.
\end{abstract}

\section{Introduction}

Digital texts are central to the study, dissemination, and preservation of human knowledge. 
Tens of thousands of users consult digital archives daily in Europe alone \citep{10.5555/3200334.3200364},
yet billions of documents remain trapped in hard copy in libraries and archives around the world. 
These documents contain extremely diverse character sets, languages, fonts or handwriting, printing technologies, and artifacts from scanning and aging.  Converting them into machine-readable data that can power indexing and search, computational textual analyses, and statistical analyses - and be more easily consumed by the public - requires highly extensible, accurate, efficient tools for optical character recognition (OCR). 

Current predominant OCR technology -- developed largely for small-scale commercial applications in high resource languages -- falls short of these requirements. OCR is typically modeled as a sequence-to-sequence (seq2seq) problem, with learned embeddings from a neural vision model taken as inputs to a learned neural language model. The seq2seq architecture is challenging to extend and customize to novel, lower resource settings \citep{hedderich-etal-2021-survey}, because training a vision-language model requires a vast collection of labeled image-text pairs and significant compute.
This study shows that on printed Japanese documents from the 1950s, the best performing existing OCR mis-predicts over half of characters. 
Poor performance is widespread, 
spurring a large post-OCR error correction literature \citep{10.1162/tacl_a_00379, 10.1145/3453476, artidigh20} and skewing digital history towards limited settings that are not representative of the diversity of documentary history.

\begin{figure*}[ht]
    \centering
    \includegraphics[width=.75\linewidth]{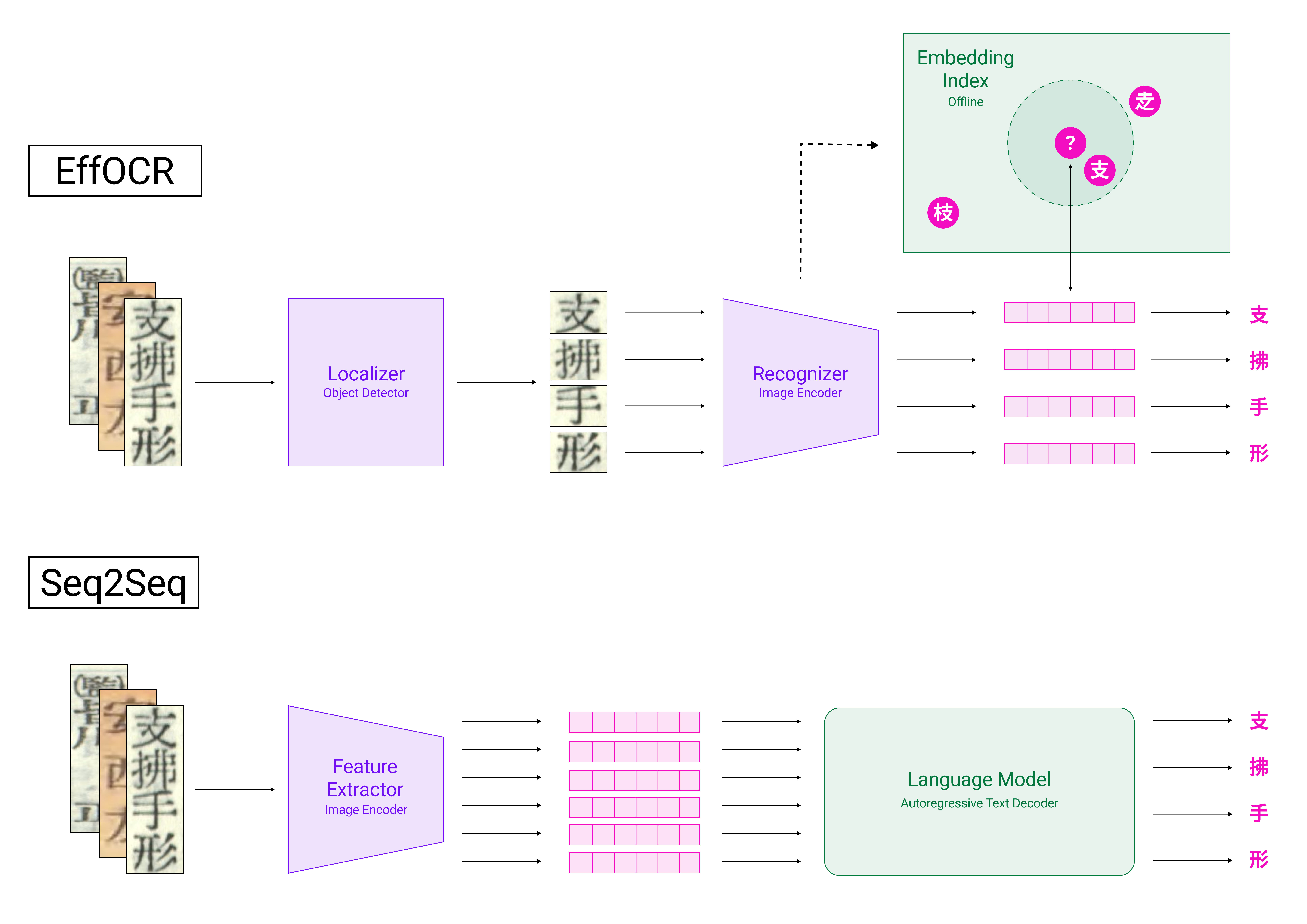}
    \caption{\textbf{EffOCR and Seq2Seq Model Architectures.} This figure represents the EffOCR architecture, as compared to a typical sequence-to-sequence OCR architecture.} 
    \label{fig:arch}
    \vspace{-4mm}
  \end{figure*}

This study develops a novel, open source OCR architecture, EffOCR (\textbf{Eff}icient\textbf{OCR}), designed for researchers and archives seeking a sample-efficient, customizable, scalable OCR solution for diverse documents. EffOCR combines the simplicity of early OCR systems, such as Tauschek's 1920s reading machine, 
with deep learning, bringing OCR back to its roots: the \emph{optical} recognition of \emph{characters}. Deep learning-based object detection methods are used to localize individual characters or words in the document image.
Character (word) recognition is modeled as an image retrieval problem, using a vision encoder contrastively trained on character (word) crops.

EffOCR performs accurately, even when using lightweight models designed for mobile phones that are cheap to train and deploy.
Using documents that are fundamental to studying Japan's remarkable 20th century economic growth, the study shows EffOCR can provide a sample efficient, highly accurate OCR architecture for contexts where all current solutions fail. 
EffOCR's blend of accuracy and efficient runtime also makes it attractive for digitizing massive-scale collections in high resource languages, which the study illustrates with Library of Congress's collection of historical U.S. newspapers \citep{locca}. EffOCR has been used to cheaply and accurately digitize the over 20 million page scans in this collection \citep{dell2023american}. 

In principle, contextual understanding could be extremely valuable to OCR, but in practice state-of-the-art transformer seq2seq models are extremely costly to train, expensive to deploy, and do not exist for lower resource languages, with advances concentrated in a handful of languages. 
This study shows that taking a step back from seq2seq models unlocks massive gains in sample efficiency. Researchers, with a modest number of annotations and modest compute, can train their own OCR for settings where all existing solutions fail, using our user-friendly EffOCR open-source package.
New characters specific to a setting can also be added at inference time -- since they don't need to be seen in sequence during training -- important for contexts such as archaeology and certain historical applications where new characters are regularly encountered. These features facilitate making digital history more representative of documentary history.

\section{Methods}

Modern OCR overwhelmingly uses deep neural networks -- either a convolutional neural network (CNN) or vision transformer (ViT) -- to encode images. The representations created by passing an input image through a neural encoder are then decoded to the associated text.

Figure \ref{fig:arch} underscores two fundamental differences between EffOCR and seq2seq. 
First, sequence-to-sequence architectures typically require line level inputs, and individual characters or words are not localized; rather, images or their representations are divided into fixed size patches. In contrast, EffOCR localizes characters and words using modern object detection methods \citep{cai2018cascade, Jocher_YOLOv5_by_Ultralytics_2020} via the ``localizer'' module.
Second, seq2seq sequentially decodes the learned image representations into text using a learned language model that takes the image representations as inputs. In contrast, EffOCR recognizes text by using contrastive training \citep{khosla2020supervised} to learn a meaningful metric space for character or word-level OCR. A vision encoder, the ``recognizer'' module, projects crops of the same character (word) -- regardless of style -- nearby, whereas crops of different characters (words) are projected further apart. 

EffOCR thus generates full lines of text in the following way: (1) the localizer produces bounding boxes for characters (words) in the input image; (2) these localized character (word) images are embedded with the recognizer; (3) the character (word) embeddings are decoded to machine-readable text in parallel by retrieving the label of their nearest neighbor in an offline index of exemplar character (word) embeddings, created by rendering labeled character (word) images with a digital font; and (4) the bounding boxes from the localizer are re-used to robustly infer the order of the machine-readable characters (words) and the presence of white spaces. Embedding distances are computed using cosine similarity with a Facebook Artificial Intelligence Similarly Search (FAISS) backend \citep{johnson2019billion}. The vision embeddings alone are sufficient to infer text since they represent characters -- not text lines like in seq2seq -- and hence decoding them does not require a language model with learned parameters.

This study develops both character and word level OCR models, with the former being more suitable for character-based languages and the latter more suitable for alphabet-based languages. 
When modeling OCR as a word level problem, EffOCR defaults to character level recognition if the distance between a word crop embedding and the nearest embedding in the offline dictionary of word embeddings is below a threshold cosine similarity. This is important, as hyphenated words at the end of lines, acronyms, proper nouns, and antiquated terms often make it infeasible to construct a comprehensive word dictionary. 

EffOCR is trained on digital font renders, along with a modest number of labeled crops from target datasets.
The recognizer is trained using the Supervised Contrastive (``SupCon'') loss function  \citep{khosla2020supervised}, a generalization of the InfoNCE loss \citep{oord2018representation} that allows for multiple positive and negative pairs for a given anchor. We use the ``outside'' SupCon loss formulation,
% omitted for clarity second equivalence: =\sum_{i \in I} \mathcal{L}_{\text {out }, i}^{\text {sup }}

\begin{small}
\[ \mathcal{L}_{\text {out }}^{\text {sup }}=\sum_{i \in I} \frac{-1}{|P(i)|} \sum_{p \in P(i)} \log \frac{\exp \left(\boldsymbol{z}_i \cdot \boldsymbol{z}_p / \tau\right)}{\sum_{a \in A(i)} \exp \left(\boldsymbol{z}_i \cdot \boldsymbol{z}_a / \tau\right)} \]
\end{small}
as implemented in PyTorch Metric Learning \citep{musgrave2020pytorch}, where $\tau$ is the temperature, $i$ indexes a sample in a ``multiviewed" batch (in this case multiple fonts/augmentations of characters with the same identity), $P(i)$ is the set of indices of all positives in the multiviewed batch that are distinct from $i$, $A(i)$ is the set of all indices excluding $i$, and $z$ is an embedding of a sample in the batch.

To create training batches for the recognizer, EffOCR uses a custom $m$ per class sampling algorithm without replacement.
This metric learning batch sampling algorithm also implements batching and training with hard negatives, where the negative samples in a batch are selected to be semantically close to one another, and thus contrasts made between anchors and hard negatives may be especially informative. %Indeed, one of the main advantages of contrastive training is that it allows the learning process to exploit hard negative mining. More details are provided in the supplementary materials.

Different vision encoders can be used interchangeably for the EffOCR character localizer - which locates the character/word crops -- and recognizer -- which learns a metric space for these crops. Three models are considered for character level EffOCR: a vision transformer model (EffOCR-T Base) with XCiT (Small) \citep{ali2021xcit} for both the localizer and recognizer, a convolutional base model (EffOCR-C Base) with ConvNeXt (Tiny) \citep{liu2022convnet} for both the localizer and recognizer, and a convolutional small model (EffOCR-C Small), which uses lightweight architectures designed for mobile phones -- YOLOv5 (Small) \citep{Jocher_YOLOv5_by_Ultralytics_2020} for the localizer and MobileNetV3 (Small) for the recognizer. For word level OCR, we develop EffOCR-Word (Small), which uses the same lightweight architectures as EffOCR-C (Small). EffOCR-Word (Small) defaults to EffOCR-C (Small) when the cosine similarity between a word crop embedding and the nearest embedding in the offline word embedding dictionary is below 0.82, a hyperparameter that is (like all model hyperparameters) tuned on the validation set. The base models use a two-stage object detector for character localization, specifically a Cascade R-CNN \citep{Cai_2019}, whereas the small models use one-stage object detection for faster speed \citep{Jocher_YOLOv5_by_Ultralytics_2020}. The supplementary materials describe the EffOCR architecture and training recipes with no detail spared and evaluate models using alternative vision transformer encoders. 

%ConvNeXt is a new state-of-the-art CNN backbone, XCiT was chosen because of its comparative advantage in modeling fine-grained features via the ability to accommodate smaller patch sizes through a linear complexity attention mechanism, which may be especially suitable for character images with small spatial extents (as measured in pixels), and MobileNetV3 (Small) and YOLOv5 (Small) were collectively chosen to produce a speed optimized EffOCR, as both architectures are popular, easily customizable, and speed-optimized by design.\footnote{The inference speed advantages offered by a smaller transformer encoder, such as MobileViT, are much more modest than that offered by MobileNetV3, and hence an EffOCR-T (small) model is not developed.}
%As the deep learning literature advances and new models are developed, EffOCR's modular framework and simple training recipes make it straightforward to swap in new encoders, granting the model a degree of future-proofness.

\section{Related Literature}

EffOCR's architecture draws inspiration from metric learning methods for efficient 
image retrieval  \citep{el2021training}, joining a recent literature on self-supervision through simple data augmentation for image encoders  \citep{grill2020bootstrap, chen2021empirical, chen2021exploring}. 
The closest frameworks to EffOCR in their overall design are the original OCR conceptualizations, such as Tauschek's 1920s reading machine, which used human engineered features to recognize localized characters.
More recently, CharNet \citep{xing2019convolutional}, developed for scene text (not documents), uses separate convolutional networks for dense classification and regression at a single scale, outputting a character class and bounding box at every spatial location, and then aggregates this information with confidence scores to make final predictions. EffOCR in contrast deploys widely used, highly optimized object detection methods to localize characters and then feeds character crops to a contrastively trained recognizer.\footnote{Others have also used contrastive learning for OCR, in particular \citep{aberdam2021sequence} use a self-supervised, sequence-to-sequence contrastive learning approach.} Other OCR frameworks - that are widely used, have state-of-the-art performance, or provide an instructive architectural contrast with EffOCR -  are described in Section \ref{compare}, which introduces the comparisons that we will make.

\section{Training and Evaluation datasets}

Evaluating EffOCR 
requires benchmark datasets that are representative of the diversity of documentary history. Traditional OCR benchmarks focus on commercial applications like receipts \citep{huang2019icdar2019} - and SOTA OCR systems evaluate on these data - which are not relevant to digital history. 

\begin{figure*}[ht]
    \centering
    \includegraphics[width=.7\linewidth]{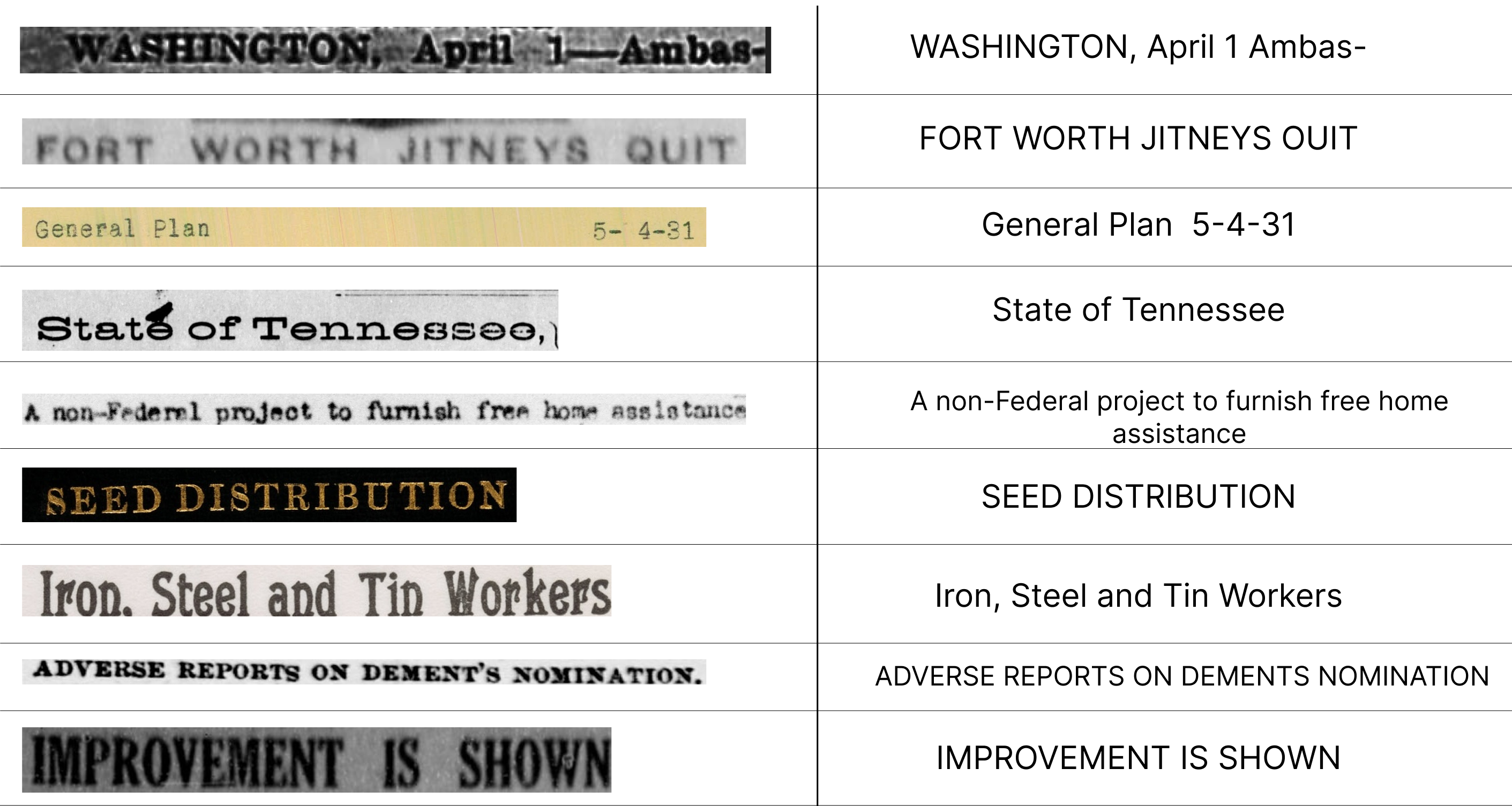}
    \caption{\textbf{Diversity in the Chronicling America Dataset.} This figure shows examples sampled from the Chronicling America (LoCCA) dataset, along with EffOCR predicted transcriptions.} 
    \label{fig:locca}
    \vspace{-4mm}
  \end{figure*}
\bigskip

Instead, the study draws on the literature on historical image datasets \citep{nikolaidou2022survey}. First, it uses documents from historical Japan that can elucidate fundamental questions that have been understudied due to a lack of digital data, such as the drivers of Japan's rapid transformation from a poor agrarian economy to a wealthy industrialized nation. Horizontally and vertically written tabular data -- providing rich information on Japanese firms and their personnel -- are drawn from two 1950s publications \citep{pr, teikoku}. A 1930s prose publication providing detailed biographies of tens of thousands of individuals  \citep{ww} is also examined.   
These texts could use over 13,000 \textit{kanji} characters. %, many only subtly different.  

The second context is Library of Congress's Chronicling America (LoCCA) collection, which contains over 19 million historical public domain newspaper page scans. This collection is highly diverse, as shown in Figure \ref{fig:locca}. 

Library of Congress provides an OCR, but the quality is low \citep{smith2015computational}. 
There is a large literature studying historical newspapers at scale, which overwhelmingly uses keyword search and does not unlock the power of large language models due to poor quality digitization \citep{hanlon2022historical}. 
LoCCA elucidates how EffOCR: 1) performs in the highest resource setting, English; 2) extensibility across Latin and \textit{kanji} characters, which differ significantly in their aspect ratios and complexity; 3) extensibility to the many Unicode renderable languages that use the Latin script.

Layout datasets exist for Chronicling America and some of the Japanese publications \citep{shen2020large,lee2020newspaper}. Adding word/character bounding boxes and transcription annotations builds upon the existing work of the historical image dataset literature \citep{nikolaidou2022survey}. 

Because seq2seq requires lines as inputs, to build the Japanese and Chronicling America datasets we draw lines at random from the Japanese volumes and from 10 randomly selected newspapers in LoCCA. 
Lines correspond to cells in tables and single lines within columns/rows in prose. 
The baseline training sets range from 291 lines %(7,438 characters) 
for Chronicling America to 1309 cells 
for horizontal Japanese, highly feasible for researchers to label in an afternoon, and also includes validation and test splits. The annotations were double-entered by the study authors, with all discrepancies hand-resolved. While the randomly selected lines/table cells in the labeled data can contain names, the underlying images are already public.

For the newspapers, we also provide an additional evaluation-only dataset that consists of a sample of 225 textlines, randomly drawn from all scans in the Chronicling America collection published on March 1st of years ending in ``6,'' from 1856-1926. This sample is balanced across these decades, with 25 textlines sampled randomly from each of the days. A selection of textlines from this set is shown in Figure \ref{fig:locca}. The day-per-decade set is designed to be challenging, by weighting older, much harder to read scans from the mid-19th century equally despite their relative scarcity in the Chronicling America collection. 

In addition to this gold quality data, we create silver quality training data for training EffOCR-Word (Small) by applying the EffOCR-C (Small) model to a random sample of newspapers. We limited the number of crops with model-generated labels to 20 -- so each word can have 0-20 silver-quality crops depending upon its frequency of occurrence in our random sample. This limit is binding for common words, \textit{e.g.,} ``the.'' We also use the gold word crops from the 291 line training set, which cover only a small share of words. Using silver quality data leads to high performance, achieved essentially for free. The study's datasets are publicly released.

Finally, we examine EffOCR on an existing Polytonic Greek benchmark \citep{gatos2015grpoly}, selected because it contains both line-level and word transcriptions. Polytonic Greek uses five diacritics to notate older Greek texts. It is challenging because the diacritics have a similar appearance.  
The supplemental materials show example documents from all benchmarks.

\section{Measurement and comparisons} \label{compare}

OCR accuracy is measured using the character error rate (CER), the Levenshtein distance between the OCR'ed string and the ground truth, normalized by the length of the ground truth. A CER of 0.5, for instance, translates to mispredicting approximately half of characters. 

The most widely used OCR engines are commercial products that do not currently support fine-tuning and have proprietary architectures. The study compares EffOCR to Google Cloud Vision (GCV) and Baidu OCR (popular for Asian languages). 
We include these comparisons because they are relevant to practitioners.

We also consider four open source architectures:
EasyOCR's convolutional recurrent neural network (CRNN) framework \citep{shi2016end}, TrOCR's sequence-to-sequence encoder-decoder transformer (base and small) \citep{li2021trocr}, Tesseract's bi-directional LSTM, and PaddleOCR's Single Vision Text Recognition (SVTR), which also abandons seq2seq, dividing text images into small (non-character) patches, using mixing blocks to perceive inter- and intra-character patterns, and recognizing text by linear prediction \citep{du2022svtr}.  A large literature has examined a variety of custom-designed OCR systems. We focus on those that either (1) make similar architectural choices (SVTR), (2) are considered SOTA, regardless of architectural choices (TrOCR), or (3) are very popular (Tesseract and EasyOCR). 

\begin{table*}[ht]
    \centering
    \resizebox{\linewidth}{!}{
    \begin{threeparttable}
       \begin{tabular}{lcccccccccccccc}
      \toprule
      &		&	&		& 	&	& \multicolumn{6}{c}{Character Error Rate} & &	\multicolumn{2}{c}{Lines/second} \\
	&		&	& 	&	&	&	Horiz.	&	Vertical 	&	Vertical	&	\multicolumn{2}{c}{Chron. Amer.}	& Anci. & &	Horiz. &	Chron. 	\\
Model/Engine	& Seq2Seq?	&	Transformer? & Pretraining	&	Parameters	& &	Jap.	&	Jap. (tables)	&	Jap. (prose)	&	Eval & Day/Decade	 & Greek & &	Jap.	&	Amer.	\\

\cmidrule{1-5} \cmidrule{7-12} \cmidrule{14-15} \\

EffOCR-C (Base)	&	$\times$ & $\times$ 	&	from scratch	&	112.5 M	&&	\underline{\textbf{0.006}}	&	\underline{\textbf{0.007}}	& 0.030 & 0.023 & 0.062 & 0.049  & & 0.79 & 0.49	\\ \\

EffOCR-C (Small)	&	$\times$ & $\times$ 	&	from scratch	&	9.3 M	&	& 0.010 & 0.009 & 0.036 & 0.028 & 0.080 & 0.052 && 19.46 & 13.40	\\ \\
EffOCR-T (Base) & $\times$ & $\checkmark$ & from scratch & 101.8 M&& 0.009 & 0.007 & \underline{\textbf{0.027}} & 0.022 & 0.059 & \underline{\textbf{0.047}} & & 0.19 & 0.31 \\ \\

EffOCR-Word (Small) &	$\times$ & $\times$ 	&	from scratch	&	10.6 M	&	& - & - & - & 0.015 & 0.043 & - && - & 21.36	\\ \\
 \\

Google Cloud Vision OCR &  ? & ?	&	off-the-shelf	&	?	& &	0.173	&	0.695	&	0.135	&	\underline{\textbf{0.005}} & \underline{\textbf{0.019}} & 0.065 & & ? & ?	\\ \\

Baidu OCR	&	? & ?	&	off-the-shelf	&	? 	& &	0.060	&	0.556	&	0.177	& -&	- & - & & ? & ?	\\ \\

Tesseract OCR (Best) 	&	$\checkmark$ & $\times$	&	off-the-shelf	& 1.4 M & &	1.021	&	0.996	&	0.744	&	0.106 & 0.170  & 0.251 & & 4.90 & 4.47	\\ \\

EasyOCR CRNN 	& $\checkmark$	&	$\times$ &	off-the-shelf	&	3.8 M	&&	0.191	&	- 	&	 -	&	0.170 & 0.274  & - & & 33.55 & 19.80 	\\
	&	&	&	fine-tuned	&		&	 &	0.082	&	 -	&	 -	&	0.036 & 0.157	\\
 
     &    &    & from scratch & 	&	&	0.132	&	 -	&	 -	&	0.131  & 0.204 &  \\ \\
   
PaddleOCR SVTR	&  $\times$	& $\times$	&	off-the-shelf	&	11 M	& &	0.085	&	 -	&	 -	&	0.304 & 0.314 & - & & 13.34 & 13.56	\\
	&	&	&	fine-tuned	&		&&	0.032	&	 -	&	- 	&	0.103 & 0.129 & \\
     &   & & from scratch &		&	&	0.097	&	- 	&	- 	&	0.104 &	0.138 & \\ \\
  
TrOCR (Base)	& $\checkmark$	& $\checkmark$ &	off-the-shelf	&	334 M	&		&	-	&	-	&	-	&	0.015 & 0.038 & - & & - & 0.43	\\
&	&		&	fine-tuned	&				&& -	&	-	&	-	&	0.013 & 0.027	\\
    &    & & from scratch &		&		&	-	&-	&	-	&	0.809 & 0.831 &	 \\ \\
 
TrOCR (Small)	& $\checkmark$  &	$\checkmark$ 	&	off-the-shelf	&	62 M	&&	- &	-	&	-	&	0.039 & 0.121 & - & & - & 0.97	\\
	&	&	&	fine-tuned	&		&		& -	&	-	&	-	&	0.075 & 0.091 &	\\
      &  & & from scratch &		&		& - &	-	&	-	&	0.773 & 0.820 &	 \\
 
    \bottomrule 
    \end{tabular}
    \end{threeparttable}}
    \caption{\textbf{Baseline Results and Comparisons.} This table reports the performance of different OCR architectures, \textit{off-the-shelf} (without fine-tuning on target data), \textit{fine-tuned} on the target publication training set from a pre-trained OCR checkpoint, and trained \textit{from scratch} on synthetic text lines and the target publication training set. ``?'' indicates that the field is unknown due to the proprietary nature of the architecture.}
        \label{comparisons_rebuttal}
\end{table*}

The pre-trained EasyOCR, PaddleOCR, and TrOCR models are fine-tuned on the same target data as EffOCR. 
Considerable resources have been devoted to pre-training these models. For example, TrOCR was pre-trained on 684 million English synthetic text lines. Hence, these comparisons elucidate performance when these pre-trained models are further tuned on the target datasets. 
For a more apples-to-apples comparison, the study examines the accuracy of these architectures when trained from scratch (using a pre-trained checkpoint not trained for OCR, when supported by the architecture) on 8,000 synthetic text lines (like EffOCR) and the same target crops. 
EasyOCR and PaddleOCR do not support vertical Japanese, and TrOCR does not support any Japanese.
Tesseract offered little support for fine-tuning until recently and hence most of its applications have been off-the-shelf, which is this study's focus. All results come from a single model run, with training details provided in the supplemental materials.

\section{Results} \label{results}
 EffOCR provides a highly accurate OCR with minimal training data, in contexts where current solutions fail.
For vertical Japanese tables, the best EffOCR CER is 0.7\% (Table \ref{comparisons_rebuttal}). The next best alternative, Baidu OCR, has a CER of 55.6\%, making nearly 80 times more errors. The best EffOCR CER is modestly higher for the Japanese prose (2.7\%); these scans are low resolution and some characters are illegible, to provide a context where OCR with language modeling could offer a clear advantage. Yet EffOCR makes 5 times fewer errors than the next best alternative (GCV), whose CER of 13.5\% will not support applications that require high accuracy.  For horizontal Japanese -- a higher resource setting -- the EffOCR CER is 0.6\%, whereas the next-best-alternative (Paddle OCR fine-tuned on target crops) makes more than five times more errors. The different EffOCR models produce strikingly similar results, despite the significant differences in architecture (convolutional versus transformer) and model size (9.3M to 112.5M parameters).
By making an accurate digitization of such collections feasible - with minimal training data requirements - EffOCR can contribute to the diversity of digital texts available to researchers. 

The CER (uncased) for the LoCCA newspapers is 1.5\%. GCV has the best performance (0.5\%), followed by fine-tuned TrOCR (Base) (1.3\% CER). The advantage of EffOCR on English - the quintessential high resource setting - is its open-source codebase and fast runtime.
GCV makes significant layout errors when fed full newspaper page scans, which have complex layouts \citep{shen2021layoutparser}, and hence the performance in Table \ref{comparisons_rebuttal} cannot be replicated when it is fed scans. GCV charges per image, and the supplementary materials estimate a cost at current prices of \$23 million USD to digitize LoCCA at the line image level, versus \$60K for EffOCR-Word (Small), which researchers have used to cheaply and accurately digitize this collection \cite{dell2023american}. 

Table \ref{comparisons_rebuttal} examines CPU runtime for open source architectures, measured by lines processed per second on identical dedicated hardware (four 2200 MHz CPU cores, selected to represent a plausible and relatively affordable research compute setup). GPUs are prohibitively costly for mass digitization. 
EffOCR-Word (Small) is 50 times faster than TrOCR (Base), which is likely to be cost prohibitive for larger scale applications. EffOCR supports inference parallelization across characters -- promoting faster inference -- whereas seq2seq requires autoregressive decoding. On English, the most plausible scalable alternative is fine-tuned EasyOCR. With a third of the parameters of EffOCR-Word (Small), it is slightly slower and the CER is around 29\% higher. 
For horizontal Japanese, EffOCR-C (Small) is three times more accurate and faster than PaddleOCR SVTR (fine-tuned), the next best alternative.

 \begin{figure*}[ht]
    \centering
    \includegraphics[width=0.9\linewidth]{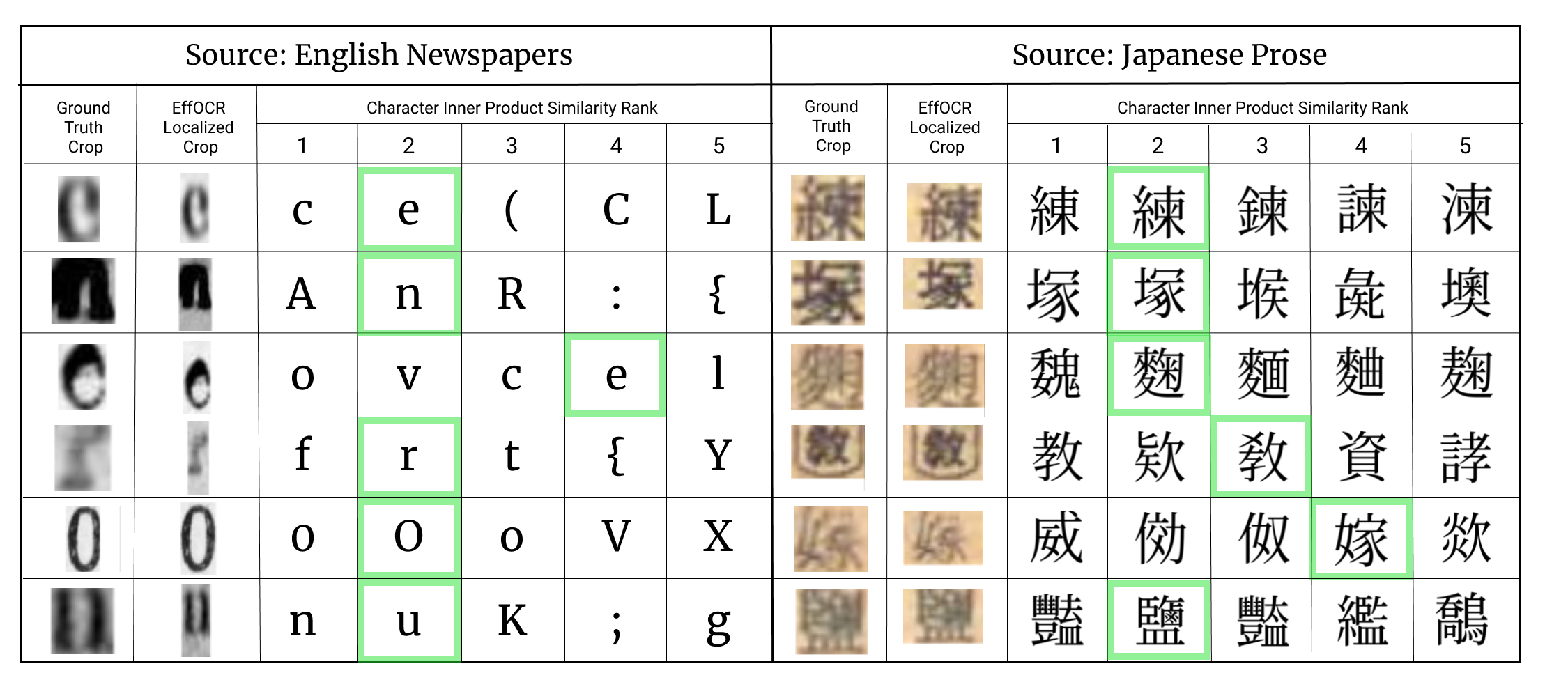}
    \caption{\textbf{Error Analysis.} Representative examples of EffOCR errors, showing the target crop, the EffOCR localized crop, and the five nearest characters in the embedding index, with the correct character highlighted in green.} 
    \label{fig:errors}
    \vspace{-4mm}
  \end{figure*}
 \bigskip
 
Figure \ref{fig:errors} provides representative examples of errors, showing the target crop, the localized crop, and its five nearest neighbors, with the correct prediction highlighted in green. 
Errors tend to occur when the character is illegible or homoglyphic to another character (\textit{e.g.,} O and 0). For example, a 0 in one font can occasionally be indistinguishable from an O in another, an error that would be straightforward to correct in post-processing. 

The supplementary materials report results from additional encoders, and examine how different architecture and design choices for EffOCR contribute to its performance. In particular, we notice little difference between the best performing CNN encoders and vision transformer encoders in terms of CER, regardless of language, when holding approximately constant the number of model parameters. This is consistent with an existing literature on the convergent performances of (appropriately modernized) CNNs and vision transformers \cite{liu2022convnet}.

EffOCR outperforms all other architectures that support Polytonic Greek, including Google Cloud Vision. This illustrates the versatility of the architecture. 

EffOCR's parsimonious architecture 
allows it to learn efficiently.
%This can be quantified with an apples-to-apples comparison, where EffOCR-C (Base) is compared to leading open source architectures pre-trained from scratch on the same number of synthetic text lines as EffOCR and tuned on the same target crops. Table \ref{comparisons_rebuttal} shows that EffOCR learns faster than both seq2seq (CRNN, TrOCR) and competing vision only (SVTR) architectures, with the next best alternative making 16 times more errors for Japanese and 4.7 times more errors for LoCCA. The transformer seq2seq model, which is extremely data hungry, is unusable.
To quantify this, we train different OCR models from scratch using varying amounts of annotated data.
All architectures are pre-trained from scratch on 8,000 synthetic text lines, starting from pre-trained checkpoints not customized for OCR when supported by the framework. They are then fine-tuned on the study's benchmark datasets, with varying train splits: 70\%, 50\%, 20\%, 5\%, and 0\% (using only synthetic data).
These exercises are performed for Chronicling America and horizontal Japanese, as vertical Japanese is not supported by the comparison architectures. 

\begin{figure*}[ht]
    \centering
    \includegraphics[width=.7\linewidth]{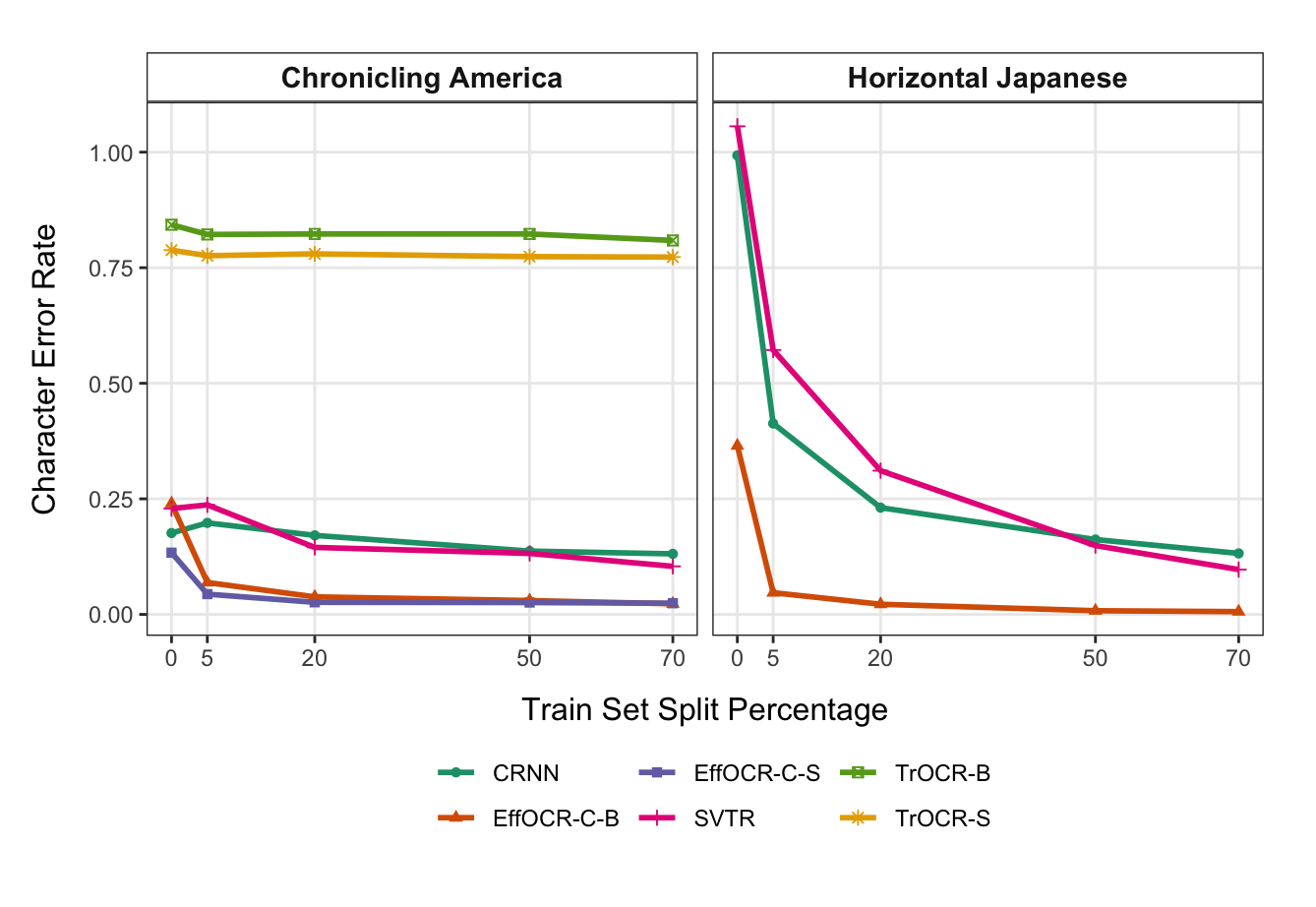}
    \caption{\textbf{Sample Efficiency.} This figure plots the percentage of the benchmark dataset used in training against the character error rate, for different OCR model architectures.} 
    \label{fig:efficiency}
    \vspace{-4mm}
  \end{figure*}

Figure \ref{fig:efficiency} plots the percentage of the benchmark dataset used in training on the x-axis and the CER on the y-axis.
On just 99 labeled table cells for Japanese and 21 labeled rows for LoCCA (the 5\% train split), EffOCR's CER is around 4\%, showing viable few shot performance.  
The other architectures remain unusable. 
EffOCR performs nearly as well using 20\% of the training data as using 70\%, where it continues to outperform all other alternatives. 

Here, our focus is on the design of bespoke, efficient models for low-resource contexts. 
One might wish to assess how EffOCR performs on completely out-of-domain texts. Elsewhere, researchers have used the EffOCR package and EffOCR-Word (Small) model trained only on newspapers to process randomly selected, highly diverse documents from the U.S. National Archives \cite{bryan-etal-2023-efficientocr}. EffOCR performs similarly to other open-source OCR engines, achieving a CER of 11.2\% as compared with a 11.8\% CER from Tesseract (Best), a 12.1\% CER from EasyOCR, and a 51\% CER from TrOCR (Small). The sample efficiency of EffOCR suggests it could be trained to perform well off-the-shelf on diverse archival documents by labeling a small number of samples across a wide range of common historical document types, an effort that could be crowd-sourced.

\section{Discussion} \label{discussion}

Indexing, analyzing, disseminating, and preserving diverse documentary history 
requires community engagement of stakeholders with the requisite fine-grained knowledge of the relevant settings. EffOCR facilitates this engagement because it is highly extensible to low-resource settings, sample-efficient to customize, and simple and cheap to train and deploy.
In contrast, seq2seq is more aligned with the commercial objective of designing a product that is difficult for competitors to imitate.  
For example, EffOCR can be trained in the cloud with free student compute credits, whereas
TrOCR required training on a multi-million dollar cluster with 32 32GB V100 cards. Lower resource languages may lack the pre-trained language models required to initialize a transformer seq2seq model, and sufficient compute resources are also unlikely to be available.  
EffOCR encourages community engagement by integrating the follow features:

\textbf{Character/word level}:
EffOCR creates semantically rich visual embeddings of individual characters (words), a parsimonious problem. Annotators can select which of the most probable predictions from the pre-trained recognizer are correct, potentially using a simple mobile interface, or line level labels can be mapped to the character (word) level once a localizer has been developed. 

\textbf{Language Extensibility}: Language modeling advances have concentrated around less than two dozen modern languages, out of many thousands \citep{joshi2020state}.
Omitting the language model makes EffOCR extensible and easy-to-train. To extend EffOCR to a new language, all one needs are renders for the appropriate character set. 
Additionally, characters do not need to be seen in sequence during training, so new characters can be added at inference time, valuable for archaeological contexts where new characters are regularly discovered.
 Omitting the language model makes it easy to mix scripts, necessary for some languages. 
The recognizer can also be exposed to characters in training using any desired sequencing. This is not true of multilingual seq2seq training, which leads to many OCR errors with endangered languages \citep{rijhwani2020ocr}.
%EffOCR can convert the Japanese publications examined in this study into a knowledge graph, revealing rich economic insights about Japanese economic development that were previously unknown due to the failure of all existing OCR solutions. The supplementary materials provide more details.

\textbf{Decoupling localization and recognition:} Theoretically, localization and recognition (akin to classification) may rely on different features of the image, suggesting modularity \citep{song2020revisiting}. Practically, decoupling allows localization and recognition to use different training sets, economizing on annotation costs since these tasks can require very different numbers of labels depending on the script. 
It also encourages community innovation and future-proofness, because it simplifies training recipes and makes it straightforward to swap in new localizers or recognizers - including zero-shot models such as \citet{kirillov2023segment} - as the literature advances. 

\textbf{Scalable:} 
The small EffOCR models achieve fast CPU inference that can scale cheaply to hundreds of millions of documents. %EffOCR-Word (Small) has been used to digitize the approximately 20 million scans in the Chronicling America collection on a \$60,000 budget for inference, whereas TrOCR would have cost nearly 50 times as much and GCV would have been orders of magnitude more expensive. 
 
\textbf{Open-Source:} The open-source EffOCR python package \cite{bryan-etal-2023-efficientocr} makes it straightforward to use existing EffOCR models off-the-shelf with just a few lines of code, including for those who lack familiarity with deep learning frameworks. It also includes functionality to train custom models and guides users with tutorials.

\section{Reproducibility} 
We release all code and training data used to create EffOCR. Scripts in the public repository exactly reproduce the figures cited above. All other material needed to reproduce these results is detailed in the supplemental materials, including training hyperparmeters. The models in this paper can also be deployed through the open-source EffOCR python package (CC-BY 4.0 license). 

%An emerging literature explores foregoing OCR altogether to directly reason on images 
%\citep{park2019cord, guo2019eaten, mathew2021docvqa, rust2022language}.
%The main focus is on commercial applications like receipts and train tickets. There are not end-to-end models for performing most research tasks, and document images are large to store.
%Hence, OCR is likely to remain relevant for academic applications and digital archives for the foreseeable future.
%EffOCR's simple character image retrieval framework can expand accessibility to a rich diversity of human knowledge.  

\section{Limitations}

%EffOCR is not currently an off-the-shelf solution. 
%Rather, it is designed for contexts where researchers need fine-grained control. 
%More extensive pre-training would be necessary to evaluate EffOCR's zero-shot capabilities.
%Its sample efficiency suggests the feasibility of designing an EffOCR that works well for many applications off-the-shelf by exposing the pre-trained model to a modest number of crowd-sourced crops from a wide range of documents. Another next step is to build upon the data augmentation and style transfer literatures \citep{alaluf2021restyle, tumanyan2022splicing} to generate more diverse synthetic pre-training data.

This study does not focus on handwriting due to space constraints, but the approach would be analogous. Synthetic handwriting generators, \textit{e.g.,} \citet{bhunia2021handwriting}, could provide extensive data for pre-training, analogous to this study's use of digital fonts.

There are some settings where EffOCR's framework is not suitable. 
If large portions of a document are illegible, context is necessary. Moreover, the heavy use of ligatures and/or slanting in some character sets and handwriting could lead to more challenging character localization. This challenge is mitigated with the word-level EffOCR model.

\section{Ethical Considerations}

EffOCR presents no major ethical concerns. Its methods are entirely open source, and its training data are entirely in the public domain. Its core functionality, accurately transcribing texts in low-resource settings, is ethically sound. By making it easier to digitize scanned document texts in low-resource settings, it can promote the inclusion of more diverse groups in NLP, social science, and humanities research. Its sample and computational efficiency minimizes environmental harm by reducing compute requirements at training and inference time. 

Some applications of EffOCR could raise ethical flags. We discourage users from applying EffOCR to copyrighted documents unless the application is protected by fair use. While EffOCR is a potentially useful tool for studying bias, \textit{e.g.,} through analyses of historical documents, potentially harmful or offensive content transcribed by EffOCR should not be shared without proper context.   

\clearpage

\appendix

\section*{Materials and Methods}
\label{sec:appendix}
\subsection*{Encoders}

Different encoders can be used interchangeably for EffOCR's character localization module (hereafter, ``localizer") and character recognizing module (hereafter ``recognizer"). We use the following:
\begin{itemize}
    \item \textbf{EffOCR-C (Base)}: ConvNeXt (Tiny) \cite{liu2022convnet} for both the localizer and recognizer. Both models are initialized from the officially released checkpoint with specifications: \\
    \texttt{\{size: "tiny"\}}
    \item \textbf{EffOCR-T (Base)}: XCiT (Small) \cite{ali2021xcit} for both the localizer and recognizer. Both models are initialized from the officially released checkpoint with specifications:  \\
    \texttt{\{size: "small", depth: 12, patch\_size: 8, resoultion: 224\}}
    \item \textbf{EffOCR-C (Small)}: YOLOv5 (Small) \cite{Jocher_YOLOv5_by_Ultralytics_2020} for the localizer and MobileNetV3 (Small) \cite{howard2019searching} for the recognizer. YOLOv5 is initialized from the officially released \texttt{YOLOv5s} checkpoint, and MobileNetV3 is initially from the PyTorch Image Models (``timm") \cite{rw2019timm} produced checkpoint with specifications: \\
    \texttt{\{size: "small", channel\_multiplier: 0.50\}}
\end{itemize}

For ablations, we also examine: 
\begin{itemize}
    \item Swin (Tiny) \cite{liu2021swin} for both the localizer and recognizer. Both models are initialized from the officially released checkpoint with specifications:\\
    \texttt{\{size: "tiny", patch\_size: 4, window: 7, resolution: 224\}}
    \item ViTDet (Base) \cite{li2022exploring} for the localizer and a vanilla vision transformer, ViT (Base), for the recognizer. Both models are initialized from the officially released checkpoint with specifications: \\
    \texttt{\{size: "base", patch\_size: 16, resolution: 224\}}
\end{itemize}

These architectures were selected for the following reasons:

\begin{itemize}
    \item  \textbf{EffOCR-C (Base)}: ConvNeXt is a new state-of-the-art CNN backbone, in contrast to the other three vision transformer encoders.
    \item \textbf{EffOCR-T (Base)}: XCiT was chosen because of its comparative advantage in modeling fine-grained features via the ability to accommodate smaller patch sizes through a linear complexity attention mechanism, which may be especially suitable for character images with small spatial extents (as measured in pixels).
    \item \textbf{EffOCR-C (Small)}: MobileNetV3 (Small) and YOLOv5 (Small) were collectively chosen to produce a speed optimized EffOCR, as both architectures are popular, easily customizable, and speed-optimized by design.
   \item The Swin transformer was selected because of its state-of-the-art performance on object detection tasks.
  \item The original ViT embeddings perform well for image retrieval, and have become a new baseline for image retrieval \cite{el2021training}.
\end{itemize}

The inference speed advantages offered by a smaller transformer encoder, such as MobileViT, are much more modest than that offered by MobileNetV3, and hence an EffOCR-T (small) model is not developed, although it would be straightforward to do so should users desire it. In tests, a MobileViTv2 (small) Recognizer model was approximately 6.5 times slower than a comparable MobileNetv3 Recognizer. 

As the deep learning literature advances and new models are developed, EffOCR's modular framework and simple training recipes make it straightforward to swap in new encoders, granting the model a degree of future-proofness.

These models are all trained on a single A6000 GPU card, with hyperparameters selected using the 15\% validation split, save for the models with XCiT (Small) or ViT (Base) encoders, which were trained on two A6000 GPU cards.

\bigskip

\subsection*{Character Localization} 
All models use an MMDetection \cite{chen2019mmdetection} backend for localization, except for the ViTDet ablation, which uses Detectron2 \cite{wu2019detectron2} and YOLOv5 (Small) \cite{Jocher_YOLOv5_by_Ultralytics_2020} for EffOCR-C (Small), which uses its own custom training scripts. 
Only one EffOCR configuration, EffOCR-C (Small), has a localizer that uses a one-stage object detection framework: YOLOv5 (Small) \cite{Jocher_YOLOv5_by_Ultralytics_2020}. All others use a two-stage object detector, specifically a Cascade R-CNN \cite{Cai_2019}.
One stage object detection is faster, and hence makes sense for the small model, where a central objective is fast inference speed. 

The localizers built with ConvNeXt (EffOCR-C Base), XCiT (EffOCR-T Base), and Swin (ablation)  are trained on 8,000 textlines of synthetic data for 40 epochs at a constant learning rate of $1e-4$ and fine-tuned on benchmark data for 100 epochs at a $2.5e-5$ constant learning rate, all with anchor generator scales $[2,8,32]$.
ViTDet is trained on 8,000 textlines of synthetic data for 40 epochs with a constant learning rate of $1e-4$, and then fine-tuned for 100 epochs on benchmark data with a $1e-5$ constant learning rate. The YOLO localizer is trained on 8,000 textlines of synthetic data for 30 epochs at a constant learning rate of $1e-2$ and fine-tuned on benchmark data for 30 additional epochs, still at a constant $1e-2$ learning rate.

%\bigskip

%\textbf{Synthetic Data:} 
The synthetic data used for pre-training the localizers and comparison models was created using a custom synthetic data generator. %, which can found at the ``EffSynth" GitHub repository \cite{cbd2023effsynth}. 

This generator was used to create six synthetic dataset variants, each consisting of 10,000 synthetic lines with an 80\%-10\%-10\% train-test-validation split. The six dataset variants are: horizontal English with character sequences generated at random, horizontal Japanese with character sequences generated at random, vertical Japanese with character sequences generated at random, horizontal English with text sequences generated from Wikipedia, horizontal Japanese with text sequences generated from (Japanese) Wikipedia, and vertical Japanese with text sequences generated from (Japanese) Wikipedia. Text sequence based synthetic datasets were used to pre-train seq2seq models that rely on language context, e.g., TrOCR and CRNN; character sequence based synthetic datasets were used to pre-train non-seq2seq models, e.g., EffOCR and SVTR. 

\bigskip

\subsection*{Character Recognition}

The EffOCR recognizer is trained using the Supervised Contrastive (``SupCon") loss function  \cite{khosla2020supervised}, a generalization of the InfoNCE loss \cite{oord2018representation} that allows for multiple positive and negative pairs for a given anchor, as described in the main text.

To create training batches for the recognizer, EffOCR uses a custom $m$ per class sampling algorithm \textit{without replacement} adapted from the PyTorch Metric Learning repository \cite{musgrave2020pytorch}.

This metric learning batch sampling algorithm also implements batching and training with hard negatives, where the negative samples in a batch are selected to be semantically close to one another, and thus contrasts made between anchors and hard negatives may be especially informative for the model to update on. Indeed, one of the main advantages of contrastive training is that it allows the learning process to exploit hard negative mining.

More specifically, the custom batch sampling algorithm samples $m$ character variants for each class (character) - drawn from both target documents and augmented digital fonts. We choose $m=4$ and the batch size is 1024, meaning 4 styles/representations of each of 256 different characters appear in each batch. The model learns to map character crops of the same identity to similar dense vectors in a semantically rich, high-dimensional vector space, and vice versa.
 For EffOCR recognizer training, an epoch is defined as some number $P$ passes through all unique characters $N$ in the character set under consideration, i.e., $N=13,738$ for Japanese and $N=91$ for English. Empirically, a good setting for Japanese is $P=1$, so the total number of classes in an epoch is 13,738, and for English $P=10$, so the total number of classes in an epoch is 910. 
Sampling for each class occurs without replacement, for better coverage of character variants. Because of this, the number of passes $P$ matters, as it determines the number of character variants used for contrastive training in each epoch.

Every character crop that appears in the training set is embedded using a model first trained without hard negative mining/sampling, and for each we find its 8 nearest neighbors.
The EffOCR recognizer is then trained again from scratch, with batches being sampled with an $m$ per class sampler (without replacement) that is further modified to randomly intersperse hard negative sets (8 nearest neighbor characters, $m=4$ variants of each) throughout batches. 

EffOCR is trained on digital font renders from readily available fonts (13 for Japanese and 14 for English), along with a modest number of labeled crops from the target datasets.\footnote{Fonts for Japanese included: Dela Gothic One Regular; Hachi Maru Pop Regular; Hina Mincho Regular; Komorebi Gothic; Kosugi Regular; New Tegomin Regular; Noto Serif CJK JP Regular; Reggae One Regular; Shippori Mincho B1 Regular; Stick Regular; taisyokatujippoi7T5; Tanugo Regular; and Yomogi Regular. Fonts for English included: Anton Regular; Cutive Mono Regular; EB Garamond Regular; Fredoka Regular; IM Fell DW Pica Regular; NewYorker-jLv; Noto Serif Regular; Oldnewspapertypes-449D; Orbitron Regular; Special Elite Regular; Ultra Regular; VT323 Regular; ZaiConsulPolishTypewriter-MVAxw; and ZaiCourierPolski1941-Yza4q.} %See the EffOCR GitHub repository for the font files themselves \cite{cbd2023effocr}. 
The digital fonts are augmented by randomly applying affine transformations (translation and scaling); background coloring, color jittering, color inversion, and grayscaling; and Gaussian blurring. The model trains on digital fonts and labeled crops \textit{together}, since the objective is to learn general purpose embeddings that would map target crops nearby to digital renders. All recognizer models except MobileNetV3 use an AdamW optimizer with weight decay of $5e-4$, a SupCon loss with temperature of 0.1, a learning rate of $2e-5$, and a batch size of 128. MobileNetv3 uses the same parameters except a learning rate of $2e-3$. The Japanese datasets are trained for 60 epochs, character-level English is trained for 30, and word level for 40 epochs.

After recognizer training is completed, the recognizer is used as an encoder to create an offline index of exemplar character embeddings to be searched at inference time for the purposes of character recognition. Specifically, the exemplar character embedding index is created by embedding image renders for all the unicode characters supported by the Google Noto Serif font series, i.e., Noto Serif CJK JP Regular for models trained for Japanese OCR and Noto Serif Regular for models trained for English OCR. The Google Noto series is chosen as an exemplar font due to both its extremely wide coverage of glyphs and the simplicity of its style, though, by virtue of EffOCR's training, other fonts could be used as well. At inference time, FAISS \cite{johnson2019billion} is used to perform an \textit{inner product} similarity search that compares character embeddings in the sample being inferenced to exemplar character embeddings in this offline index; identities are assigned to inferenced characters using the identity of that character's nearest neighbor in the offline exemplar index, i.e., k-NN classification with $k=1$.

For case sensitive applications, EffOCR character recognition for English text can also be lightly post-processed to help better differentiate uppercase and lowercase letters from one another: one can force a character to be uppercased or lowercased through simple rules based statistics about the dimensions of bounding boxes (in the sample undergoing inference). This procedure is irrelevant for results reported in this text, however, for which CER is measured uncased. 

Checkpoints/weights for all recognizers are supported by implementations from timm \cite{rw2019timm}.

\subsection*{Comparisons} 

To examine sample efficiency, we train alternative architectures from scratch, on the same number of synthetic text lines used to train EffOCR. 
Specifically, the comparison architectures are, as applicable, initialized with ``default" pre-trained checkpoints that have not yet been exposed to an OCR task, e.g., masked language model pre-trained weights for text transformers or ImageNet pre-trained weights for CNNs and vision transformers. These comparison architectures are then trained on 8,000 synthetic text lines per the applicable synthetic dataset variant (see: Methods - Synthetic Data) as a form of standardized OCR-task-specific pre-training. They are then fine-tuned on the same benchmark datasets used to assess EffOCR, but with varying train-test-validation splits: 70\%-15\%-15\%, 50\%-25\%-25\%, 20\%-40\%-40\%, 5\%-47.5\%-47.5\%, and 0\%-50\%-50\% (i.e., zero-shot).

The hyperparameters used for initializing and training comparison models are as follows: 

\begin{itemize}
    \item The EasyOCR implemented \textbf{CRNN} \cite{shi2016end} comparison is trained from a random initialization (as is the default in EasyOCR) for 100,000 iterations on the horizontal English text sequence and horizontal Japanese text sequence synthetic datasets, respectively. The learning rate is fixed at 1.0 with an Adadelta optimizer and the batch size is 128, per the EasyOCR configuration defaults. The architecture uses VGG for feature extraction, a BiLSTM for seq2seq/language modeling, and a CTC loss, as also is the EasyOCR default. A new prediction head is used to match the character set associated with EffOCR for Japanese. The resulting model is then fine-tuned for 30,000 iterations with a batch size of 64, and all other hyperparameters the same, on the benchmark datasets of varying splits.
    \item The \textbf{SVTR} \cite{du2022svtr} comparison is first trained from a random initialization for 500 epochs with an Adam optimizer with cosine-scheduled learning rate of 0.001 and batch size of 32 on horizontal English character sequence and horizontal Japanese character sequence synthetic datasets, respectively. All these hyperparameters are PaddleOCR defaults, which are also used for fine-tuning on the benchmark dataset splits.
    \item The \textbf{TrOCR} \cite{li2021trocr} comparison models are initialized from the appropriate vision transformer and language transformer pre-trained encoder and decoder checkpoints: for TrOCR (Base) this is the officially released BEiT (Base) checkpoint and the officially released RoBERTa (Large) checkpoint used by the TrOCR authors for model initialization; for TrOCR (Small) these are similarly the officially released checkpoints for DeiT (Small) and MiniLM used by the TrOCR authors for their model initialization. These checkpoints are exported directly from the TrOCR GitHub repository \cite{trocr2023Github} using a modified script originally authored by Hugging Face \cite{wolf-etal-2020-transformers}, such that training is possible in native PyTorch with Huggingface model implementations. TrOCR (Base) is trained on the horizontal English synthetic text sequence dataset for 60 epochs at a fixed learning rate of $5e-7$ with a batch size of 16; TrOCR (Small) is trained for 40 epochs, with all other hyperparameters the same. (The learning rate was selected based on experiments with the validation set.) The resulting models are then fine-tuned with the same hyperparameters on the various benchmark dataset splits.
\end{itemize}

To evaluate how existing solutions perform when fine-tuned on the EffOCR benchmark datasets, existing pre-trained checkpoints from the EasyOCR CRNN, PaddleOCR SVTR, and TrOCR (Base) and TrOCR (Small) models are fine-tuned on the baseline 70\%-15\%-15\% split of the benchmark datasets. Specifically, the 15\% validation set is used for hyperparameter tuning and the 15\% test set is used to construct the results reported in the study. 

For all comparison models, training hyperparamters are the same as used during the sample efficiency assessments with standardized synthetic pre-training, save that prediction heads for relevant models are left as they are by default. Model initialization differs, accordingly: TrOCR (Base) and TrOCR (Small) use \texttt{microsoft/trocr-base-stage1} and \\ \texttt{microsoft/trocr-small-stage1} checkpoints, respectively; EasyOCR CRNN uses the most recently released \texttt{japanese\_g2.pth} and \texttt{english\_g2.pth} checkpoints; and PaddleOCR SVTR uses the most recently released \texttt{japan\_PP-OCRv3\_rec\_train} and \\\texttt{en\_PP-OCRv3\_rec\_train} best accuracy checkpoints.

\bigskip

\subsection*{Inference Speed Comparisons}
For digitizing large-scale collections, fast inference on a CPU is necessary, due to the high costs of GPU compute. All comparisons are made on four 2200 MHz CPU cores, selected to represent a plausible and relatively affordable research compute setup. To standardize measurements of speed, each model generated predictions on the same 15\% test set.
All EffOCR models are implemented with ONNX Runtime for cross-compatibility and speed. 

Inference speed is inherently dependent on implementation and it is plausible that the other open-source architectures may be updated in the future to achieve faster inference speeds. A strong correlation between model size and inference speed is apparent and intuitive, highlighting the utility of the EffOCR-C (Small) model for digitizing knowledge - like the Chronicling America collection - at scale. 
 
A random sample of 10 LoCCA scans shows an average of 1944 column x lines per scan (historical newspapers used small fonts and contained few images), which implies the cost at current prices to digitize the LoCCA collection at the line level using GCV would be over 23 million US dollars. 

Using FS4 VM instances in Microsoft Azure to process all content in the LoCCA collection for one randomly selected day per decade, on average it took 17.21 seconds to process 1,000 lines with EffOCR-C (small). At current prices, this translates to a cost of  \$0.000908 per one thousand lines, as compared to GCV's current prices of \$1.50 (first 5 million units) and \$0.60 (above 5 million units) per thousand lines to process Chronicling America at the line level. 

\subsection*{Benchmark Dataset Creation}

Figure \ref{fig:datasets} illustrates the documents used to create this study's benchmarks. 
The OCR systems evaluated in this study take lines (cells in tables or individual lines from columns in prose) as inputs.
These segments were created using a Mask R-CNN \cite{he2017mask} model custom-trained with Layout Parser \cite{shen2021layoutparser}, an open-source package that provides a unified, deep learning powered toolkit for recognizing document layouts. 
Mask R-CNN was applied to the three Japanese publications considered and to ten different newspapers randomly selected from Chronicling America. 
Segments were selected at random for inclusion in this study's benchmark datasets. Table \ref{dataset} provides dataset statistics.

To create the character region and text annotations, three highly skilled annotators annotated each segment. All discrepancies were then hand checked and resolved. 
Each of the datasets has a 70\%-15\%-15\% train-validate-test split used for baseline evaluations.
The validation set was used for model development, whereas the test set was used only once, to create the results reported in this study. 

To create the silver quality data used to train EffOCR-Word (small), we apply the EffOCR-C (Small) model to a random sample of days. We limited the number of crops with model-generated labels to 20 - so each word can have 0-20 silver-quality crops depending upon its frequency of occurrence in our random sample. This limit is binding for common words, \textit{e.g.,} "the".

\clearpage

\section*{Data Sheet}

% Color picked from the Datasets for Datasheets paper
\definecolor{darkblue}{RGB}{46,25, 110}

\newcommand{\dssectionheader}[1]{%
   \noindent\framebox[\columnwidth]{%
      {\fontfamily{phv}\selectfont \textbf{\textcolor{darkblue}{#1}}}
   }
}

\newcommand{\dsquestion}[1]{%
    {\noindent \fontfamily{phv}\selectfont \textcolor{darkblue}{\textbf{#1}}}
}

\newcommand{\dsquestionex}[2]{%
    {\noindent \fontfamily{phv}\selectfont \textcolor{darkblue}{\textbf{#1} #2}}
}

\newcommand{\dsanswer}[1]{%
   {\noindent #1 \medskip}
}

%\begin{singlespace}
%\begin{multicols}{2}

%%%%%%%%%%%%%%%%%%%%%%%%%%%%%%%%%%%%%%%%%%

\dssectionheader{Motivation}

\dsquestionex{For what purpose was the dataset created?}{Was there a specific task in mind? Was there a specific gap that needed to be filled? Please provide a description.}

\dsanswer{ The dataset was created to train EfficientOCR (EffOCR) models. EffOCR models require two types of data:
\begin{enumerate}
    \item Text-line character and word localization examples
    \item Labeled character and word images, where the label is the character or word represented in the crop.
\end{enumerate} 
}

\dsquestion{Who created this dataset (e.g., which team, research group) and on behalf of which entity (e.g., company, institution, organization)?}

\dsanswer{ Anonymity Period. 
}

\dsquestionex{Who funded the creation of the dataset?}{If there is an associated grant, please provide the name of the grantor and the grant name and number.}

\dsanswer{ Anonymity Period.
}

\dsquestion{Any other comments?}

\dsanswer{ None. }

%%%%%%%%%%%%%%%%%%%%%%%%%%%%%%%%%%%%%%%%%%
\bigskip
\dssectionheader{Composition}

\dsquestionex{What do the instances that comprise the dataset represent (e.g., documents, photos, people, countries)?}{ Are there multiple types of instances (e.g., movies, users, and ratings; people and interactions between them; nodes and edges)? Please provide a description.}

\dsanswer{   Individual instances fall into ones of three categories:
\begin{enumerate}
    \item Character and word localization examples. An image of a line of text with accompanying annotations for word and character bounding boxes. 
    \item Labeled character crops. A cropped character image with accompanying character label.
    \item Labeled word crops. A cropped word image wiht accompanying word label.
\end{enumerate}

}

\dsquestion{ How many instances are there in total (of each type, if appropriate)?}

\dsanswer{ There are five distinct data sources represented in the dataset:
\begin{enumerate}
    \item Chronicling America (horizontal English text lines): 417 labeled text lines, 2313 labeled words, 10873 labeled characters. 
    \item Japanese Personnel Records (horizontal Japanese tables): 1870 labeled text lines, 4444 labeled characters. 
    \item Teikoku (vertical Japanese tables): 1283 labeled text lines, 4674 labeled characters.
    \item Who's Who (vertical Japanese prose): 657 labeled text lines, 8006 labeled characters.
    \item Polytonic Greek (horizontal Ancient Greek text lines): 652 labeled text lines, 36846 labeled characters.
\end{enumerate}
}

\dsquestionex{Does the dataset contain all possible instances or is it a sample (not necessarily random) of instances from a larger set?}{ If the dataset is a sample, then what is the larger set? Is the sample representative of the larger set (e.g., geographic coverage)? If so, please describe how this representativeness was validated/verified. If it is not representative of the larger set, please describe why not (e.g., to cover a more diverse range of instances, because instances were withheld or unavailable).}

\dsanswer{  The dataset is a sample from larger sets of textlines. In the case of Chronciling America, samples were taken uniformly across space (publication location) and time (publication date) of newspapers. In the case of Japanese records, sampling was conducted randomly accross pages of the larger works. Greek records constitute the entire labeled dataset provided. }

\dsquestionex{What data does each instance consist of? “Raw” data (e.g., unprocessed text or images) or features?}{In either case, please provide a description.}

\dsanswer{ Each character and word label include an image and a textual label. Labels are provided in the image's file name. Localization examples are in COCO format. Each collection of images is accompanied by a json file with bounding box labels. 
}

\dsquestionex{Is there a label or target associated with each instance?}{If so, please provide a description.}

\dsanswer{ Yes. Text localization examples' labels are COCO format bounding box labels. Character crop labels are unicode characters represented in the image. Word crop labels are unicode words represnted in the image. 
}

\dsquestionex{Is any information missing from individual instances?}{If so, please provide a description, explaining why this information is missing (e.g., because it was unavailable). This does not include intentionally removed information, but might include, e.g., redacted text.}

\dsanswer{ No informtation from the samples (described earlier) is missing. 
}

\dsquestionex{Are relationships between individual instances made explicit (e.g., users’ movie ratings, social network links)?}{If so, please describe how these relationships are made explicit.}

\dsanswer{ The only relationships between instances are labeled character and word crops originating from the same text lines. Each recognition example includes a unique image identifier that can provide relationships. 
}

\dsquestionex{Are there recommended data splits (e.g., training, development/validation, testing)?}{If so, please provide a description of these splits, explaining the rationale behind them.}

\dsanswer{Yes. In addition to the 'all.json' file, each collection includes prepared splits for several train/val/test splits. These are labeled 'trainXX.json', 'testXX.json', where XX is the percentage of a dataset included in the split. These exact splits were used for all benchmarking described in the paper. 
}

\dsquestionex{Are there any errors, sources of noise, or redundancies in the dataset?}{If so, please provide a description.}

\dsanswer{All characters were labeled by the researchers. Character crops were double-labeled. Some errors may remain, but this is unlikely. }

\dsquestionex{Is the dataset self-contained, or does it link to or otherwise rely on external resources (e.g., websites, tweets, other datasets)?}{If it links to or relies on external resources, a) are there guarantees that they will exist, and remain constant, over time; b) are there official archival versions of the complete dataset (i.e., including the external resources as they existed at the time the dataset was created); c) are there any restrictions (e.g., licenses, fees) associated with any of the external resources that might apply to a future user? Please provide descriptions of all external resources and any restrictions associated with them, as well as links or other access points, as appropriate.}

\dsanswer{The data is self-contained.}

\dsquestionex{Does the dataset contain data that might be considered confidential (e.g., data that is protected by legal privilege or by doctor-patient confidentiality, data that includes the content of individuals non-public communications)?}{If so, please provide a description.}

\dsanswer{The dataset does not contain information that might be viewed as confidential.}

\dsquestionex{Does the dataset contain data that, if viewed directly, might be offensive, insulting, threatening, or might otherwise cause anxiety?}{If so, please describe why.}

\dsanswer{The dataset does not contain content in any of these categories.}

\dsquestionex{Does the dataset relate to people?}{If not, you may skip the remaining questions in this section.}

% We recommend taking a broad interpretation of whether a dataset relates to people. For example, any dataset containing text that was written by people relates to people.

\dsanswer{The dataset relates to people in that it contains text written by people. However, no piece of text is more than one line long, making the text largely independent from its authors. }

\dsquestionex{Does the dataset identify any subpopulations (e.g., by age, gender)?}{If so, please describe how these subpopulations are identified and provide a description of their respective distributions within the dataset.}

\dsanswer{The dataset does not identify any subpopulations.}

\dsquestionex{Is it possible to identify individuals (i.e., one or more natural persons), either directly or indirectly (i.e., in combination with other data) from the dataset?}{If so, please describe how.}

\dsanswer{No individuals can be identified from this dataset. }

\dsquestionex{Does the dataset contain data that might be considered sensitive in any way (e.g., data that reveals racial or ethnic origins, sexual orientations, religious beliefs, political opinions or union memberships, or locations; financial or health data; biometric or genetic data; forms of government identification, such as social security numbers; criminal history)?}{If so, please provide a description.}

\dsanswer{No.}

\dsquestion{Any other comments?}

\dsanswer{None.}

%%%%%%%%%%%%%%%%%%%%%%%%%%%%%%%%%%%%%%%%%%
\bigskip
\dssectionheader{Collection Process}

\dsquestionex{How was the data associated with each instance acquired?}{Was the data directly observable (e.g., raw text, movie ratings), reported by subjects (e.g., survey responses), or indirectly inferred/derived from other data (e.g., part-of-speech tags, model-based guesses for age or language)? If data was reported by subjects or indirectly inferred/derived from other data, was the data validated/verified? If so, please describe how.}

\dsanswer{ Text images were directly observable from document scans. Labels were created by the researchers. 
}

\dsquestionex{What mechanisms or procedures were used to collect the data (e.g., hardware apparatus or sensor, manual human curation, software program, software API)?}{How were these mechanisms or procedures validated?}

\dsanswer{ Humans labeled the words, characters, and textlines in the dataset using Label Studio. 
}

\dsquestion{If the dataset is a sample from a larger set, what was the sampling strategy (e.g., deterministic, probabilistic with specific sampling probabilities)?}

\dsanswer{ Samples were taken randomly with uniform probability from all documents in each collection. 
}

\dsquestion{Who was involved in the data collection process (e.g., students, crowdworkers, contractors) and how were they compensated (e.g., how much were crowdworkers paid)?}

\dsanswer{ The researchers labeled each instance of data and were compensated. 
}

\dsquestionex{Were any ethical review processes conducted (e.g., by an institutional review board)?}{If so, please provide a description of these review processes, including the outcomes, as well as a link or other access point to any supporting documentation.}

\dsanswer{ No.
}

\dsquestionex{Does the dataset relate to people?}{If not, you may skip the remaining questions in this section.}

\dsanswer{ Not in this sense. 
}

\dsquestion{Did you collect the data from the individuals in question directly, or obtain it via third parties or other sources (e.g., websites)?}

\dsanswer{
}

\dsquestionex{Were the individuals in question notified about the data collection?}{If so, please describe (or show with screenshots or other information) how notice was provided, and provide a link or other access point to, or otherwise reproduce, the exact language of the notification itself.}

\dsanswer{
}

\dsquestionex{Did the individuals in question consent to the collection and use of their data?}{If so, please describe (or show with screenshots or other information) how consent was requested and provided, and provide a link or other access point to, or otherwise reproduce, the exact language to which the individuals consented.}

\dsanswer{
}

\dsquestionex{If consent was obtained, were the consenting individuals provided with a mechanism to revoke their consent in the future or for certain uses?}{If so, please provide a description, as well as a link or other access point to the mechanism (if appropriate).}

\dsanswer{
}

\dsquestionex{Has an analysis of the potential impact of the dataset and its use on data subjects (e.g., a data protection impact analysis) been conducted?}{If so, please provide a description of this analysis, including the outcomes, as well as a link or other access point to any supporting documentation.}

\dsanswer{
}

\dsquestion{Any other comments?}

\dsanswer{
}

%%%%%%%%%%%%%%%%%%%%%%%%%%%%%%%%%%%%%%%%%%
\bigskip
\dssectionheader{Preprocessing/cleaning/labeling}

\dsquestionex{Was any preprocessing/cleaning/labeling of the data done (e.g., discretization or bucketing, tokenization, part-of-speech tagging, SIFT feature extraction, removal of instances, processing of missing values)?}{If so, please provide a description. If not, you may skip the remainder of the questions in this section.}

\dsanswer{ The data was not preprocessed. We provide raw image crops and labels exactly as they were provided to EffOCR for training. 
}

\dsquestionex{Was the “raw” data saved in addition to the preprocessed/cleaned/labeled data (e.g., to support unanticipated future uses)?}{If so, please provide a link or other access point to the “raw” data.}

\dsanswer{
}

\dsquestionex{Is the software used to preprocess/clean/label the instances available?}{If so, please provide a link or other access point.}

\dsanswer{
}

\dsquestion{Any other comments?}

\dsanswer{
}

%%%%%%%%%%%%%%%%%%%%%%%%%%%%%%%%%%%%%%%%%%
\bigskip
\dssectionheader{Uses}

\dsquestionex{Has the dataset been used for any tasks already?}{If so, please provide a description.}

\dsanswer{ The dataset was used to train EffOCR models for English, Japanese, and Greek. Training procedures and results are described at length to the accompanying paper. 
}

\dsquestionex{Is there a repository that links to any or all papers or systems that use the dataset?}{If so, please provide a link or other access point.}

\dsanswer{ 
}

\dsquestion{What (other) tasks could the dataset be used for?}

\dsanswer{ These data could be used for training other OCR systems, or any other application requiring images of text and its transcriptions. 
}

\dsquestionex{Is there anything about the composition of the dataset or the way it was collected and preprocessed/cleaned/labeled that might impact future uses?}{For example, is there anything that a future user might need to know to avoid uses that could result in unfair treatment of individuals or groups (e.g., stereotyping, quality of service issues) or other undesirable harms (e.g., financial harms, legal risks) If so, please provide a description. Is there anything a future user could do to mitigate these undesirable harms?}

\dsanswer{ There are no considerations around these issues. 
}

\dsquestionex{Are there tasks for which the dataset should not be used?}{If so, please provide a description.}

\dsanswer{ No.
}

\dsquestion{Any other comments?}

%%%%%%%%%%%%%%%%%%%%%%%%%%%%%%%%%%%%%%%%%%
\bigskip
\dssectionheader{Distribution}

\dsquestionex{Will the dataset be distributed to third parties outside of the entity (e.g., company, institution, organization) on behalf of which the dataset was created?}{If so, please provide a description.}

\dsanswer{Yes. The dataset is available for public use.}

\dsquestionex{How will the dataset will be distributed (e.g., tarball on website, API, GitHub)}{Does the dataset have a digital object identifier (DOI)?}

\dsanswer{The dataset is distributed under a Creative Commons CC-BY license. The terms of this license can be viewed at \url{https://creativecommons.org/licenses/by/2.0/} The dataset is available on Huggingface. }

\dsquestion{When will the dataset be distributed?}

\dsanswer{ The dataset is currently available to the public. 
}

\dsquestionex{Will the dataset be distributed under a copyright or other intellectual property (IP) license, and/or under applicable terms of use (ToU)?}{If so, please describe this license and/or ToU, and provide a link or other access point to, or otherwise reproduce, any relevant licensing terms or ToU, as well as any fees associated with these restrictions.}

\dsanswer{ No.
}

\dsquestionex{Have any third parties imposed IP-based or other restrictions on the data associated with the instances?}{If so, please describe these restrictions, and provide a link or other access point to, or otherwise reproduce, any relevant licensing terms, as well as any fees associated with these restrictions.}

\dsanswer{There are no third party IP-based or other restrictions on the data. All source documents are in the public domain. }

\dsquestionex{Do any export controls or other regulatory restrictions apply to the dataset or to individual instances?}{If so, please describe these restrictions, and provide a link or other access point to, or otherwise reproduce, any supporting documentation.}

\dsanswer{No export controls or other regulatory restrictions apply to the dataset or to individual instances.}

\dsquestion{Any other comments?}

\dsanswer{None.}

%%%%%%%%%%%%%%%%%%%%%%%%%%%%%%%%%%%%%%%%%%
\bigskip
\dssectionheader{Maintenance}

\dsquestion{Who will be supporting/hosting/ maintaining the dataset?}

\dsanswer{ The dataset is in its final state. Huggingface hosts the dataset. 
}

\dsquestion{How can the owner/curator/ manager of the dataset be contacted (e.g., email address)?}

\dsanswer{Anonymity period.}

\dsquestionex{Is there an erratum?}{If so, please provide a link or other access point.}

\dsanswer{ No.
}

\dsquestionex{Will the dataset be updated (e.g., to correct labeling errors, add new instances, delete instances)?}{If so, please describe how often, by whom, and how updates will be communicated to users (e.g., mailing list, GitHub)?}

\dsanswer{ The dataset will be updated if labeling errors are found. Users finding errors should email the authors. 
}

\dsquestionex{If the dataset relates to people, are there applicable limits on the retention of the data associated with the instances (e.g., were individuals in question told that their data would be retained for a fixed period of time and then deleted)?}{If so, please describe these limits and explain how they will be enforced.}

\dsanswer{There are no applicable limits on the retention of data.}

\dsquestionex{Will older versions of the dataset continue to be supported/hosted/maintained?}{If so, please describe how. If not, please describe how its obsolescence will be communicated to users.}

\dsanswer{ If the dataset is updated due to errors, old versions will still be available via Huggingface. 
}

\dsquestionex{If others want to extend/augment/ build on/contribute to the dataset, is there a mechanism for them to do so?}{If so, please provide a description. Will these contributions be validated/verified? If so, please describe how. If not, why not? Is there a process for communicating/distributing these contributions to other users? If so, please provide a description.}

\dsanswer{ Others may download the dataset and add to it locally. There is no mechanism to add to the hosted version of the dataset. 
}

\dsquestion{Any other comments?}

\dsanswer{None.}

%\end{multicols}
%\end{singlespace}

\clearpage

\section*{Supplementary Results}

\subsection*{Ablations}
To elucidate which components of EffOCR are essential for its performance, several ablations are examined in Table \ref{ablations}: using a simple feedforward neural network classifier head for recognition instead of performing k-nearest neighbors classification\footnote{Implicitly, retrieving the nearest neighbor character from an index of offline exemplar character embeddings, as the EffOCR recognizer does by default, is k-NN classification with $k=1$.}, training with and without hard negatives, disabling training on synthetic data for the recognizer and localizer, and the use of alternative vision encoders. All ablations use a fixed set of hyperparameters that are associated with a specific localizer-recognizer configuration; these hyperparameters are outlined in the sections on Character Localization and Character Recognition.

Modeling character-level classification as an image retrieval problem weakly dominates the classification performance when using a standard multilayer perceptron with softmax procedure for classification. 
OCR as retrieval is chosen as the baseline not only due to its performance, but because it also allows for adding new characters at inference time (just embed a new exemplar character and add it to the offline index) - common in historical and archaeological settings - and because efficient similarity search technologies like FAISS \cite{johnson2019billion} provide fast inference.

Removing hard negatives increases the character error rate substantially, particularly for Japanese, which has many characters with highly similar visual appearances, e.g., some multi-stroke kanji are nearly identical to one another and differ only in the slants of some strokes. Using hard negatives in constrastive training effectively incentivizes the model to distinguish between these very visually similar characters. 

Training on only labels from the target documents leads to a large deterioration in performance for Japanese. This is as expected, given that only a fraction of \textit{kanji} characters appear in the small training datasets. The deterioration in performance is modest for English, where there are far fewer characters. 
The opposite is true for character localization. Localization for English is a harder problem than for Japanese because character silhouettes and aspect ratios are more variable. 

Two additional vision transformer encoders are explored: Swin (Tiny) \cite{liu2021swin} for both the localizer and recognizer and ViTDet (Base) \cite{li2022exploring} for the localizer and a vanilla vision transformer, ViT (Base), for the recognizer. The performance is similar to the base EffOCR-C and EffOCR-T models. 

\section*{Replication Materials}

As part of this submission, we provide a standalone codebase with scripts for running EffOCR training and inference. In addition, we provide the training data that was used to train the EffOCR line detector, localizer, and recognizer for the Library of Congress Chronicling America dataset.

Submission file size limitations prevent the inclusion of training and evaluation data for all benchmark datasets considered in this submission, though all such training and evaluation data is publicly available, and would be provided if not for the anonymity guidelines.

\clearpage

\setcounter{table}{0}
\renewcommand{\thetable}{S-\arabic{table}} % Setting the table number output to letters 
\setcounter{figure}{0}
\renewcommand{\thefigure}{S-\arabic{figure}} % Setting the figure number output to letters 

\begin{table*}[ht]
    \centering
   % \resizebox{\linewidth}{!}{
    \footnotesize{\begin{threeparttable}
       \begin{tabular}{lcccc}
      \toprule
	&	\textbf{Horiz. Jap. Tables}	&	\textbf{Vert. Japanese Tables}	&	\textbf{Vert. Jap. Prose}	&	\textbf{Chronicling America}	\\
\midrule
Train Lines	&	1309	&	898	&	459	&	291	\\
Val Lines	&	280	&	192	&	98	&	62	\\
Test Lines	&	281	&	193	&	100	&	64	\\
Total	&	1870	&	1283	&	657	&	417	\\
\midrule
Train Chars	&	3089	&	3296	&	5832	&	7438	\\
Val Chars	&	673	&	677	&	1063	&	1708	\\
Test Chars	&	682	&	701	&	1111	&	1727	\\
Total	&	4444	&	4674	&	8006	&	10873	\\
    \bottomrule 
    \end{tabular}
    \end{threeparttable}
    }
    \caption{This table reports the number of annotated lines and characters in the training, validation, and test sets of this study's four benchmarks.}
        \label{dataset}
\end{table*}

\begin{table*}[ht]
    \centering
    \resizebox{\linewidth}{!}{
    \begin{threeparttable}
       \begin{tabular}{lcccccccccccc}
      \toprule
	& &		&	& Feed Forward &	&	Hard Neg. & & \multicolumn{2}{c}{No Synthetic Data} &	& \multicolumn{2}{c}{Encoder}	\\
	& &	EffOCR-C (Base)	& &	Neural Net &	&	Off	& &	Recognizer	&	Localizer &  & Swin (Tiny) & ViT (Base) \\
 \cmidrule{1-1} \cmidrule{3-3} \cmidrule{5-5} \cmidrule{7-7} \cmidrule{9-10} \cmidrule{12-13}
Horizontal Japanese	& &	\textbf{0.006}	& &	0.006 &	&	0.041 &	&	0.594	&	0.009 & & 0.009 & 0.010	\\
Vertical Japanese (tables) &	&	\textbf{0.007}	& &	0.010 &	&	0.087	& &	0.700	&	0.016 & & 0.016  & 0.010 	\\
Vertical Japanese (prose) &	&	0.030	& &	0.038	& &	0.076 &	&	0.788	&	0.032  & & 0.036 & \textbf{0.027} 	\\
Chronicling America	& &	\textbf{0.023}	& &	0.037	& &	0.045	& &	0.027	&	0.068 & & 0.025  & 0.037	\\
      \bottomrule 
    \end{tabular}
    \end{threeparttable}
   }
    \caption{This table provides the character error rate. \textit{Feed Forward Neural Net} models the recognizer as a classification problem with a feed forward neural network, \textit{Hard Neg. Off} does not include hard negatives in recognizer training, \textit{No Synthetic Data} turns off synthetic data training in the recognizer and localizer, respectively, and \textit{Swin (Tiny)} and \textit{ViT (Base)} are alternative vision encoders.}
      \label{ablations}
\end{table*}

\begin{figure*}[ht]
    \centering
    \includegraphics[width=.8\linewidth]{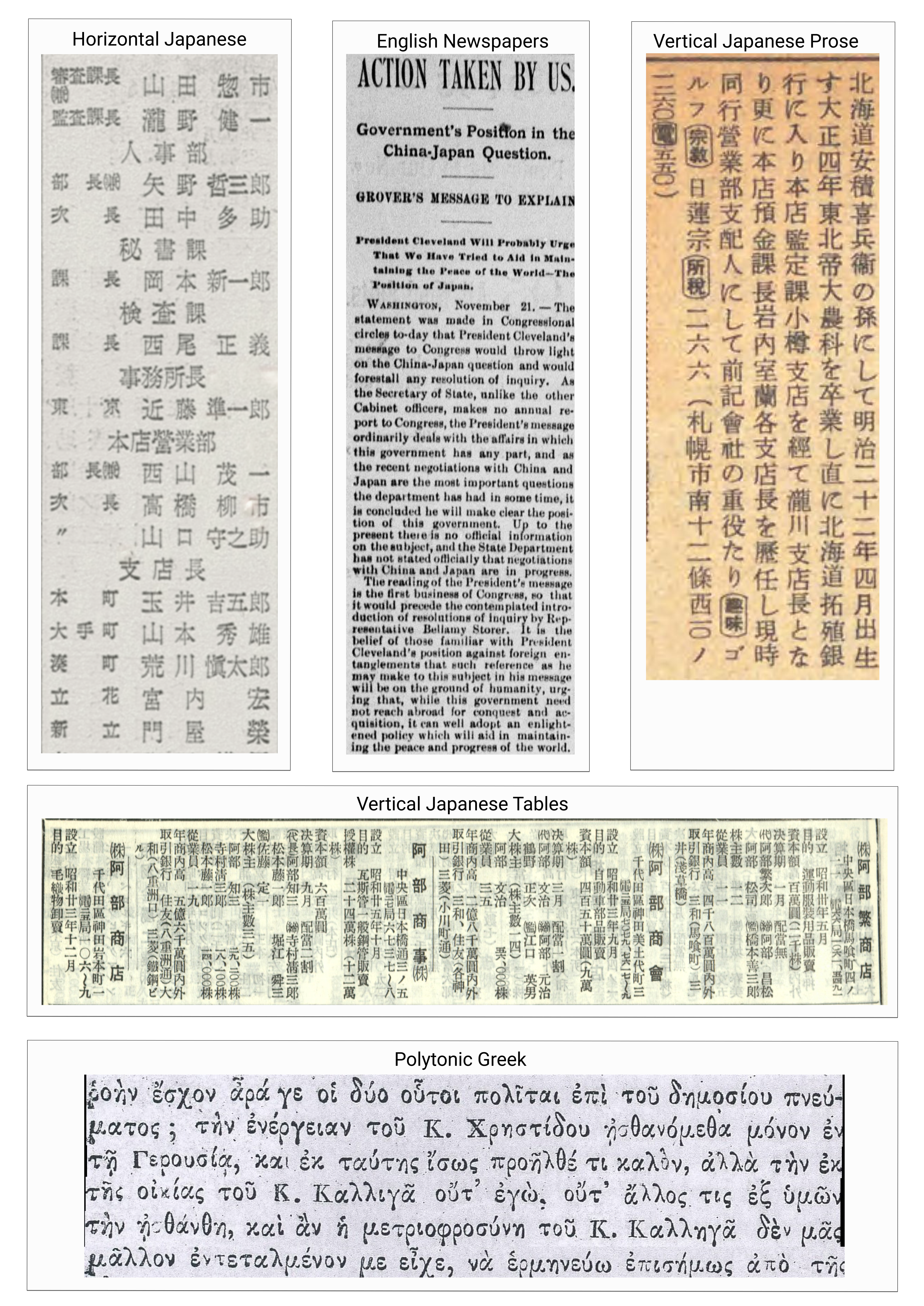}
    \caption{\raggedright \textbf{Dataset Description.} Representative samples of the publications examined in this study.} 
    \label{fig:datasets}
    \vspace{-4mm}
  \end{figure*}

\clearpage

%\textbf{Funding:} Harvard Data Science Initiative. National Science Foundation Award 1823616 

% Bibliography entries for the entire Anthology, followed by custom entries
%\bibliography{anthology,custom}
% Custom bibliography entries only
\bibliography{custom}

\end{document}